\documentclass[letterpaper,12pt,peerreviewca,draftcls]{IEEEtran}
\usepackage{csm16}
\usepackage[margin=1in]{geometry}
\usepackage{amsmath} % for \eqref
\usepackage{bm}
\usepackage{dsfont}
\usepackage{algorithmic}
\usepackage{algorithm}
\usepackage[compress]{cite}

\usepackage{url}
\usepackage{graphicx,xcolor}
\usepackage{verbatim}% http://ctan.org/pkg/verbatim
\makeatletter
\newcommand{\verbatimfont}[1]{\def\verbatim@font{#1}}%
\makeatother
\verbatimfont{\ttfamily\small}

\newcommand{\bi}{\begin{itemize}}\newcommand{\ei}{\end{itemize}}
\newcommand{\be}{\begin{equation}}\newcommand{\ee}{\end{equation}}
\newcommand{\bee}{\begin{enumerate}}\newcommand{\eee}{\end{enumerate}}
\newcommand{\bea}{\begin{eqnarray}}\newcommand{\eea}{\end{eqnarray}}
\newcommand{\beas}{\begin{eqnarray*}}\newcommand{\eeas}{\end{eqnarray*}}
\newcommand{\bc}{\begin{center}}\newcommand{\ec}{\end{center}}
\usepackage[english]{babel}
\usepackage{blindtext}
\usepackage{float}
\title{Improving the Efficiency of Gradient Descent Algorithms Applied to Optimization Problems with Dynamical Constraints\\
}
\author{Ion Matei, Maksym Zhenirovskyy, Johan de Kleer and John Maxwell\\
	POC: Ion Matei (imatei@parc.com)\\ \today }

\newif\ifPDF \ifx\pdfoutput\undefined\PDFfalse \else\ifnum\pdfoutput > 0\PDFtrue \else\PDFfalse \fi \fi
\ifPDF
\usepackage[pdftex, plainpages = false, colorlinks=true, linkcolor=black, citecolor = green!50!blue, urlcolor = blue, filecolor=black, pagebackref=false, hypertexnames=false,  pdfpagelabels ]{hyperref}
\fi

\begin{document}
\maketitle
%\CSMsetup
%\linenumbers \modulolinenumbers[2] % added for CSMAG only

\begin{abstract}
We introduce two block coordinate descent algorithms for solving optimization problems with ordinary differential equations (ODEs) as dynamical constraints. The algorithms do not need to implement direct or adjoint sensitivity analysis methods to evaluate loss function gradients. They results from reformulation of the original problem as an equivalent optimization problem with equality constraints. The algorithms naturally follow from steps aimed at recovering the gradient-decent algorithm based on ODE solvers that explicitly account for sensitivity of the ODE solution. In our first proposed algorithm we avoid explicitly solving the ODE by integrating the ODE solver as a sequence of implicit constraints. In our second algorithm, we use an ODE solver to reset the ODE solution, but no direct are adjoint sensitivity analysis methods are used.
Both algorithm accepts mini-batch implementations and show significant efficiency benefits from GPU-based parallelization. We demonstrate the performance of the algorithms when applied to learning the parameters of the Cucker-Smale model. The algorithms are compared with gradient descent algorithms based on ODE solvers endowed with sensitivity analysis capabilities, for various number of state size, using Pytorch and Jax implementations. The experimental results demonstrate that the proposed algorithms are at least 4x faster than the Pytorch implementations, and at least 16x faster than Jax implementations. For large versions of the Cucker-Smale model, the Jax implementation is thousands of times faster than the sensitivity analysis-based implementation. In addition, our algorithms generate more accurate results both on training and test data. Such gains in computational efficiency is paramount for algorithms that implement real time parameter estimations, such as diagnosis algorithms.
\end{abstract}

%\section{Preliminary Remarks}
%https://mec560sbu.github.io/2016/09/30/direct_collocation/
%http://implicit-layers-tutorial.org/neural_odes/

\section{Introduction}
Various engineering applications in system design,  control, or diagnosis are formulated as optimization problems with dynamical constraints. The most common mathematical models to describe system dynamics include ordinary differential equations (ODEs), differential algebraic equations (DAEs), or partial differential equations (PDEs). Such models rarely have closed-form solutions, leading to numerical approximation of their solutions via solvers.
Computing gradients of loss functions requires using sensitivity analysis techniques to determine the changes in the model dynamics as the optimization parameters are varied. Sensitivity analysis is performed using direct \cite{DICKINSON1976123} or adjoint \cite{CAO2002171} methods. In direct methods, the sensitivity of the solution of an ODE $\dot{x}=f(x;\theta)$, where $x$ and $\theta$ are the state vector and the model parameters, respectively,  is explicitly computed by solving an associated ODE defined in terms of the Jacobian of the map $f(x;\theta)$. The direct method works well when the number of optimization parameters $\theta$ is small, but does poorly as the size of $\theta$ increases. In contrast, using the adjoint method there is no need to explicitly compute the sensitivity of the state $x$ with respect to $\theta$. The gradient of the loss function is computed by solving the adjoint ODE that scales linearly with the size of the state vector.

Recently, deep learning platforms such Pytorch \cite{paszke2017automatic} or Jax \cite{jax2018github} were endowed with capability to include ODEs in their models. Computing gradients of loss functions over ODE solutions requires the implementation of sensitivity analysis methods (direct or adjoint) in the computational structure of these platforms. Regardless of the approach, as the complexity of these dynamical models grow so does the computational cost to solve them. Moreover, the time required to solve ODEs depends on the type of solver used (e.g., fixed/adaptime step, explicit/implicit) and on the numerical stability of the ODE. The latter is determined by the parameters explored during the optimization process and, at time, can destabilize the main ODE, or the ODE needed to be solved in the direct or adjoint methods.
In this paper we introduce algorithms that remove the need to solve the ODE induced by using direct or adjoint methods to account for the state sensitivities. This way, our algorithm are faster, since at most we have to solve only the main ODE. We are particularly interested in how the dimension of the state vector affects the computationally complexity of the optimization process. The reason for such an interest is that in applications such as model-based diagnosis we often use single or double fault assumptions that require estimating a small number of parameter compared the state dimension.

We first formulate the optimization problem with dynamical constraints. Let $x\in \mathds{R}^n$ denote a state vector that satisfies an ODE $\dot{x}=f(x;\theta)$,
where $\theta\in \mathds{R}^p$ is a set of parameters of the ODE that also serve as optimization parameters for the learning problem. Let $y=h(x)\in \mathds{R}^m$ be a vector representing indirect observations of the state vector $x$.
We denote by $\hat{\bm{x}}(\theta) = [\hat{x}_1(\theta),\hat{x}_2(\theta), \ldots, \hat{x}_N(\theta)]$ and by
$\hat{\bm{y}} = H(\hat{\bm{x}})= [\hat{y}_1,\hat{y}_2, \ldots, \hat{y}_N]$ the solution of the ODE and its corresponding outputs,
respectively, at time samples $\mathcal{T} = \{t_i\}_{i=0}^N$, with $h=t_{i+1}-t_{i}$ %and for the parameter vector $\theta$.
Let $L(\hat{\bm{y}}(\hat{\bm{x}}(\theta)))=(L\circ \hat{\bm{y}} \circ \bm{x})(\theta)=\bar{L}(\theta)$ be a scalar loss function (e.g., mean square error loss function). Then an optimization problem with dynamical constraints can be expressed as
\begin{eqnarray}
\label{eq:07182001}
  \min_{\theta} & & \bar{L}(\theta)\\
  \nonumber \label{eq:07182008}
  \textmd{such that:} & & \hat{\bm{x}}(\theta) = \textmd{{\tt ODESolver}}(\theta;x_0),\\
  \nonumber \label{eq:07182009}
                & & \hat{\bm{y}} = H(\hat{\bm{x}}(\theta)),
\end{eqnarray}
where $x_0$ is the initial condition of the ODE, and {\tt ODESolver} is an ODE solver that generates the solution of the ODE at time instants $\mathcal{T}$. A first order gradient descent algorithm to solve (\ref{eq:07182001}) is given by
\begin{equation}\label{eq:07201657}
  \theta_{k+1} = \theta_k -\alpha \nabla_{\theta}\bar{L}(\theta_k),
\end{equation}
where $\nabla_{\theta}\bar{L}(\theta)$ is the gradient of the loss function with respect to the vector of parameters $\theta$. The loss function gradient can be explicitly written as:
$$\nabla_{\theta}\bar{L}(\theta_k) = \left[\frac{\partial \hat{\bm{x}}}{\partial \theta}(\theta_k)\right]^T\left[\frac{\partial \hat{\bm{y}}}{\partial \hat{\bm{x}}}(\hat{\bm{x}}_k)\right]^T\nabla_{\hat{\bm{y}}}L(\hat{\bm{y}}_k),$$
where $\frac{\partial \hat{\bm{x}}}{\partial \theta}$ and $\frac{\partial \hat{\bm{y}}}{\partial \hat{\bm{x}}}$ are the Jacobian of the states with respect to $\theta$, and of the outputs with respect to the states, respectively, and where $\hat{\bm{x}}_k = \hat{\bm{x}}(\theta_k)$ and $\hat{\bm{y}}_k = H( \hat{\bm{x}}_k)$. We summarize the iteration (\ref{eq:07201657}) in {Algorithm \ref{alg0}}.
\begin{algorithm}
\caption{Gradient descent with sensitivities-enabled ODE solver in the loop.}
\label{alg0}
\begin{algorithmic}
\REQUIRE $\alpha$: Stepsize
\REQUIRE $x_0$: Initial state vector\\
\REQUIRE $\theta_0$: Initial parameter vector\\
        $k \leftarrow 0$
\WHILE{$\theta_k$ not converge}
\STATE $k\leftarrow k+1$
\STATE ${\theta}_k \leftarrow {\theta}_{k-1} -\alpha \nabla_{\theta}\bar{L}(\theta_{k-1})$
\ENDWHILE
\RETURN $\theta_k$
\end{algorithmic}
\end{algorithm}

In direct sensitivity analysis methods, key to the implementation of the gradient descent algorithm is the evaluation of the Jacobian of the state vector with respect to $\theta$. This quantity reflects the sensitivity of the state with respect to the vector of parameters. To avoid using numerical approximations when evaluating $\frac{\partial \bm{x}}{\partial \theta}$, ODE solvers were augmented with capability to compute sensitivities of the solution $x(t)$ with respect to $\theta$. For example, the SUNDIALS  software family \cite{gardner2022sundials,hindmarsh2005sundials}, with DAE solvers such as CVODES and IDAS, include both direct and adjoint-based  approaches to compute sensitivities. When improving a model to account for behavior that is not captured in an initial model,  one approach is to hybridize the ODE \cite{https://doi.org/10.1002/aic.690381003,MELESHKOVA2021314} by augmenting it with new representations such as neural networks (NNs). Training the parameters of the NN would require the implementation of the learning problem on a deep learning (DL) platform such as Pytorch \cite{paszke2017automatic}, Jax \cite{jaxopt_implicit_diff}, or TensorFlow \cite{tensorflow2015-whitepaper}. Currently, the algorithms in the SUNDIALS library are not directly integrated with DL platforms. Such platforms are of interest, since the optimization algorithm can take advantage of the automatic differentiation (AD) feature when computing gradients of loss functions. Recently, DL platforms were endowed with capabilities that include ODE-based layers. For example, \cite{NEURIPS2018_69386f6b} used the adjoint method to compute the gradient of the loss function, by extending the original ODE with an additional ODEs representing the adjoint variable dynamics. Similar efforts were made to extend Jax with ODE simulation capabilities \cite{optax2020github}. Regardless of the direct or adjoint approach to account for state sensitivities, as the size of the state vector increases, the evaluation of the loss function gradient will be more costly: at least linear in the state dimension, even when discounting the sometimes unpredictable numerical stability of the ODE solvers.

We introduce two block coordinate descent \cite{Bertsekas_99} algorithms that do not require implementation of sensitivity analysis method to compute loss function gradients. They are based on implicit constraints derived from the ODE and can be easily extended to dynamical models represented as DAEs. More importantly, they support batch executions over time samples that can be efficiently executed on GPUs.
%We have not observed the same time efficiency gains when using the sensitivities-enabled ODE solvers proposed in \cite{NEURIPS2018_69386f6b}.
For a vector $\bm{x} = \bm{x}_{0:N} =[x_0; x_1;x_2\ldots ;x_N]\in \mathds{R}^{n(N+1)}$, $\bm{x}_{1:N}$ is the vector $\bm{x}$ without $x_0$, and $\bm{x}_{0:N-1}$ is the vector $\bm{x}$ without the last entry $x_N$. For a function $f(x)$, $f(\bm{x}_{0:N}) = [f(x_0);f(x_1)\ldots;f(x_N)]$. We introduce the residual function ${r}(\bm{x},\theta):\mathds{R}^{n(N+1)+p}\rightarrow \mathds{R}^{nN}$ derived from the application of \emph{direct collocation} methods. An example of such a method is the Hermite-Simpson derivative collocation with trapezoid quadrature \cite{doi:10.2514/3.20223} that is given by
\begin{equation}\label{eq:07202115}
{r}(\bm{x},\theta) = -\bm{x}_{1:N} + \bm{x}_{0:N-1} + \frac{h}{6}\left[f(\bm{x}_{0:N-1};\theta)+f(\bm{x}_{1:N};\theta)+4f(\bm{x}_c;\theta)\right],
\end{equation}
where $\bm{x}_c = \frac{1}{2}\left[\bm{x}_{1:N}+\bm{x}_{0:N-1}\right]+\frac{h}{6}\left[f(\bm{x}_{0:N-1};\theta)-f(\bm{x}_{1:N};\theta)\right]$. The residual function is zero when evaluated at a solution of the ODE. In addition to the residual function, we define the loss function $F(\bm{x},\theta) = \tilde{L}(\bm{x})+\frac{1}{2}\|r(\bm{x},\theta)\|^2$, where $\tilde{L}(\bm{x}) = (L\circ \bm{y})(\bm{x})$.

\begin{algorithm}
\caption{Alternating gradient descent with residual functions based on implicit constraints.}
\label{alg1}
\begin{algorithmic}
\REQUIRE $\alpha_x$, $\alpha_{\theta}$: Stepsizes
\REQUIRE $r(\bm{x},\theta)$: Residual function as implicit dynamical constraints
\REQUIRE $x_0$: Initial state vector\\
\REQUIRE $\theta_0$: Initial parameter vector\\
        $\bm{x}_0 \leftarrow \textmd{\tt ODESolver}(\theta_0;x_0)$\\
        $k \leftarrow 0$
\WHILE{$\theta_k$, $\bm{x}_k$ not converge}
\STATE $k\leftarrow k+1$
\STATE $\bm{x}_k \leftarrow \bm{x}_{k-1} -\alpha_x \nabla_{\bm{x}}F(\bm{x}_{k-1},\theta_{k-1})$
\STATE $\theta_k \leftarrow \theta_{k-1} -\alpha_{\theta} \nabla_{\theta}F(\bm{x}_k,\theta_{k-1})$
\ENDWHILE
\RETURN $\bm{x}_k$,  $\theta_k$
\end{algorithmic}
\end{algorithm}

\begin{algorithm}
\caption{Alternating gradient descent with residual functions based on implicit constraints and state reset.}
\label{alg2}
\begin{algorithmic}
\REQUIRE $\alpha_x$, $\alpha_{\theta}$: Stepsizes
\REQUIRE $r(\bm{x},\theta)$: Residual function as implicit dynamical constraints
\REQUIRE $x_0$: Initial state vector\\
\REQUIRE $\theta_0$: Initial parameter vector\\
        $k \leftarrow 0$
\WHILE{$\theta_k$, $\bm{x}_k$ not converge}
\STATE $k\leftarrow k+1$
\STATE $\bm{x}_{k-1}\leftarrow \textmd{\tt ODESolver}(\theta_{k-1};x_0)$
\STATE $\bm{x}_k \leftarrow \bm{x}_{k-1} -\alpha_x \nabla_{\bm{x}}F(\bm{x}_{k-1},\theta_{k-1})$
\STATE $\theta_k \leftarrow \theta_{k-1} -\alpha_{\theta} \nabla_{\theta}F(\bm{x}_k,\theta_{k-1})$
\ENDWHILE
\RETURN $\bm{x}_k$,  $\theta_k$
\end{algorithmic}
\end{algorithm}
The algorithms are described in  {Algorithms \ref{alg1}} and {\ref{alg2}}.  {\tt ODESolver} is a solver that does not compute the sensitivities of the state with respect to the parameters $\theta$, thus, it is faster. In the next section, we will demonstrate that {Algorithm \ref{alg2}} is an approximation of Algorithm \ref{alg0}, where the difference comes from a redefinition of the residual function.  Both algorithms follow naturally from a gradient descent algorithm applied to the optimization problem (\ref{eq:07182001}), reformulated to explicitly include equality constraints.
 %This reformulation will not affect the optimal solution.
%Indeed, assuming that the initial value $\theta_0$ of {Algorithm \ref{alg0}} is sufficiently close to a local minimizer $\theta^*$, using the iterations introduced in {Algorithms \ref{alg1}} and  {\ref{alg2}} we are guarantee to converge to the local minima $(\bm{x}^*, \theta^*)$, with $\bm{x}^* = \textmd {\tt ODESolver}(\theta^*;x_0)$.
The vector of optimization variables $\bm{x}_k$  generated by {Algorithm \ref{alg1}} does not have to be an ODE solution except when it has converged to the local minimizer.
% However, even though it may take a different path towards the local minimizer, in the end it may still be faster than both {Algorithm \ref{alg0}} and {Algorithm \ref{alg2}} since it does not require an ODE solver in the loop.
By using the residual functions (\ref{eq:07202115}), in effect, we embed ODE solvers within the learning problem, without the need to explicitly solve ODEs. Such a formulation scales much better with the number of time samples, since we no longer have to solve, in a sequential manner, for the ODE solution and the state sensitivities.

\textbf{Accuracy of direct collocation method}:
Direct collocation methods minimize the error between the learned state trajectory and the actual ODE solution.
%Thus, the learned state trajectory is an approximation of the ODE solution.
Such local collocation  methods are both computationally simple and efficient, and support batch executions.
The collocation used in {Algorithms \ref{alg1}} and {\ref{alg2}} uses a third order polynomial to represent
the ODE solution between two time instances. The resulting approximation error can be made smaller be choosing a finer
discretization stepsize $h$. Alternatively, we can use higher order polynomials, e.g.,  fifth-order Gauss-Lobatto method
\cite{doi:10.2514/3.21662}. Such a method, based on Gauss-Lobatto quadrature (also known as Radau quadrature) is closely related
to how ODE solvers compute solutions using the implicit Radau numerical solver \cite{HAIRER199993}.
In other words, rather than separately solving the ODE, we encode the ODE solver within the optimization problem via quadrature
induced equality constraints, and jointly solve for both the parameter vector
$\theta$ and the ODE solution. %Modern ODE solvers can employ variables stepsizes when discretizing the time interval. The same idea can be used when generating the equality constraints for the residual function, by adjusting both the mesh size corresponding to time discretization and the
%degree of the approximating polynomials [ref]. Based on the error tolerance, mesh density may be reduced by merging neighboring mesh intervals or by lowering the degree of the approximating polynomial.
One advantage of the ODE solvers introduced in \cite{NEURIPS2018_69386f6b} is the access to adaptive-step solvers. Residuals in Algorithm \ref{alg2} can be formulated with an adaptive  step, as well. Indeed, since we solve an ODE at each step, the solution can provide the step adaptive quadrature coefficients that can be used to construct the residual function. For example, the {\tt scipy.integrate.solve\_ivp} function can return an {\tt OdeSolution} object that includes the time instants between which local interpolants are defined, and the local interpolants. These local interpolants can be used to construct residual functions and evaluate them at predetermined time instants.

\textbf{Memory and time efficiency}: %Similar to \cite{NEURIPS2018_69386f6b}
The memory cost is a function of the number of optimization variables (states and parameters) and the number of the residuals imposing ODE solution constraints. Gradient evaluation depends only on the computational graph of $f(x;\theta)$. %For a neural-ODE, the map $f(x;\theta)$ is a neural network, and the gradient evaluation would require performing backtracking using the computational graph of the neural network, only.
Unlike Algorithm \ref{alg0}, since there is no explicit dependence between the state variables (they are seen as independent optimization variables), state time causality has no explicit impact.  The time complexity of Algorithms \ref{alg1} and {\ref{alg2} depends on the rate of convergence of the gradient descent algorithm and the time per iteration. The rate of convergence for various versions of the gradient descent algorithm is discussed in \cite{j.2018on}. The cost per iteration (epoch) plays a big factor in the total time to generate a solution. Algorithm \ref{alg1} does not require solving ODEs, thus is the fastest. If the size of the state vector dominates the number of model parameters, the computational costs of Algorithm \ref{alg0} and \ref{alg2}, when implementing the sensitivity analysis via a direct method, are comparable, especially when implemented on GPUs.

\textbf{Mini-batch executions}:
%When using sensitivity-enabled ODE solvers, mini-batch executions can be defined with respect to initial state vectors, but it still requires solving the ODE for the entire time interval. However,
Algorithms \ref{alg1} and \ref{alg2} can define mini-batches based on time intervals: given a time window $\Delta$, we can randomly select a time instant $t_i$ and update the optimization variables based on a cost function defined using the optimization variables included in the mini-batch, only. In other words, the loss function is defined based on the state variables \{${x}_i, \ldots, {x}_{i+\Delta}$\}. The only requirement is to have the cost function $L$ separable in terms of time instants, requirement satisfied by a typical loss function such as the mean square error. In the case of Algorithm  \ref{alg2}, we can either solve the ODE for each epoch using the initial condition $x_0$, or we can reset the state by computing the solution of the ODE on the time interval corresponding to the mini-batch, where in initial state is chosen as ${x}_i$, the current estimate for the state at time instant $t_i$.

\textbf{Loss function gradient}:
In the case of Algorithm \ref{alg0}, the gradient of the loss function depends on the sensitivity of the state with respect of the optimization variables $x_{\theta} = \frac{\partial x}{\partial \theta}$. In turn, the state sensitivity $x_{\theta}$ evolves according to the ODE $\dot{x}_{\theta} = \frac{\partial f}{\partial x}x_{\theta}+\frac{\partial{f}}{\partial{\theta}}$. Therefore, the stability of the gradient of the loss function depends on the Jacobian $\frac{\partial f}{\partial x}$.  A poor choice of initial parameters $\theta$ can make the ODE unstable, leading to gradient explosion if the stepsize of the gradient descent algorithm is too small. Alternatively, a very stable ODE combined with a small $\frac{\partial f}{\partial \theta}$ can lead the fast gradient decay, hence a slow convergence of the gradient descent algorithm. Since both Algorithm \ref{alg1} and \ref{alg2} do not use explicitly the state sensitivity in the evaluation of the loss functions, their evolution will not be as significantly impacted by the dynamics of the state sensitivity.

\textbf{Experiments}:
We tested {Algorithms \ref{alg0}}-{\ref{alg2}} on the problem of learning the parameters of the particle-based Cucker-Smale ODE model \cite{doi:10.1137/090757290}.
This model is nonlinear, and we can easily increase the state vector size by increasing the number of particles. We will compare the accuracy of the algorithms both on training and testing data, and the time per iteration (epoch). We will also present loss function results based on the ODE solutions for parameter values explored during the search process.
 Such a loss function is more meaningful for comparison with {Algorithm \ref{alg0}} since {Algorithm \ref{alg1}} does not necessarily generates ODE solutions during the optimization process.

\section{Algorithms}
%
%The first order, gradient descent algorithm used to solve (\ref{eq:07182011}) takes the form
%\begin{equation}\label{eq:07191255}
%\theta_{k+1} = \theta_{k} - \alpha \nabla_{\theta}L(\theta_k),
%\end{equation}
%where $k$ denote the iteration number, $\alpha$ represents the step-size, and $\nabla_{\theta}L(\theta_k) = \left[\frac{\partial \mathcal{S}_T}{\partial \theta}\right]^T\nabla_{\bm{x}}L(\bm{x}(\theta))$.
Problem (\ref{eq:07182001}) can be reformulated as an optimization problem with equality constraints of the form
\begin{eqnarray}
\label{eq:071921301}
  \min_{\theta, \bm{x}} & & \tilde{L}(\bm{x}) \\
  \nonumber
  \textmd{such that:} & & \tilde{r}(\bm{x},\theta) = 0,
\end{eqnarray}
where $\tilde{L} = L\circ \bm{y}$,  $\tilde{r}(\bm{x},\theta) = \bm{x} - \hat{\bm{x}}(\theta)$ is the vector-valued residual function that quantifies how far $\bm{x}$ is from the solution of the ODE $\hat{\bm{x}}(\theta)=\textmd{\tt ODESolver}(\theta;x_0)$. Assuming that $\tilde{L}(\bm{x})$ attains minimum value at zero, (\ref{eq:071921301}) is equivalent to the unconstrained optimization problem
\begin{equation}\label{eq:07191302}
\min_{\theta, \bm{x}} \tilde{L}(\bm{x})+\frac{1}{2}\|\tilde{r}(\bm{x},\theta)\|^2.
\end{equation}
We later describe how we can deal with the case where $\tilde{L}(\bm{x})$ does not attain its minimum as zero. The unconstrained optimization problem (\ref{eq:07191302}) can be solved using a gradient descent algorithm, where the iterative equations are given by
\begin{eqnarray}
\label{eq:07191536}
  \bm{x}_{k+1} &=& \bm{x}_k - \alpha_{\bm{x}} \left[\nabla_{\bm{x}}\tilde{L}(\bm{x}_k) + \left(\frac{\partial \tilde{r}}{\partial \bm{x}}(\bm{x}_k, \theta_k)\right)^T\tilde{r}(\bm{x}_k,\theta_k)\right],\\
  \label{eq:07191537}
 \theta_{k+1} &=& \theta_k - \alpha_{\theta} \left[\left(\frac{\partial \tilde{r}}{\partial \theta}(\bm{x}_k, \theta_k)\right)^T\tilde{r}(\bm{x}_k,\theta_k)\right].
\end{eqnarray}
where $\nabla_{\bm{x}}\tilde{L}(\bm{x}_k) = \left(\frac{\partial \bm{y}}{\partial \bm{x}}(\bm{x}_k)\right)^T\nabla_{\bm{y}} L(\bm{y}(\bm{x}_k))$. The Jacobians of the residual function $\tilde{r}$ can be explicitly written as:
$$\frac{\partial \tilde{r}}{\partial \bm{x}} = I,\ \frac{\partial \tilde{r}}{\partial \theta} = - \frac{\partial \hat{\bm{x}}}{\partial \theta},$$
where $I$ is the identity matrix.

%-\left(\frac{\partial \mathcal{S}_T}{\partial \theta}(\theta_k)\right)^T(\bm{x}_k - \mathcal{S}_T(\theta_k))
In the following, we recover Algorithm \ref{alg0} by manipulating iterations (\ref{eq:07191536})-(\ref{eq:07191537}). We make a first modification in (\ref{eq:07191537}) to bring it closer to the iteration (\ref{eq:07201657}). Namely, we replace $\bm{x}_{k}$ by $\bm{x}_{k+1}$  resulting in
\begin{eqnarray}
\label{eq:07191543}
  \bm{x}_{k+1} &=& \bm{x}_k - \alpha_{\bm{x}} \left[\nabla_{\bm{x}}\tilde{L}(\bm{x}_k) + \left(\frac{\partial \tilde{r}}{\partial \bm{x}}(\bm{x}_k, \theta_k)\right)^T\tilde{r}(\bm{x}_k,\theta_k)\right],\\
  \label{eq:07191544}
 \theta_{k+1} &=& \theta_k - \alpha_{\theta} \left[\left(\frac{\partial \tilde{r}}{\partial \theta}(\bm{x}_{k+1}, \theta_k)\right)^T \tilde{r}(\bm{x}_{k+1},\theta_k)\right].
\end{eqnarray}
By replacing the residual function $\tilde{r}$ with the $r$ defined in terms of implicit constraints, we have obtained {Algorithm \ref{alg1}}, which can be interpreted as iteration steps in a coordinated descent method \cite{Bertsekas_99}. This change is an intermediate step to get to {Algorithm \ref{alg2}}.
%However, as we will demonstrate in the experimental results section, in itself achieves a better convergence rate as compared to iterations (\ref{eq:07191536}) - (\ref{eq:07191537}).
After this change, $\bm{x}_k$ is not guaranteed to be a solution of the ODE. We can enforce this by resetting the state at each iteration to a solution of the ODE, i.e., $\bm{x}_k = \hat{\bm{x}}(\theta_k)$. This enforcement can be done with any ODE solver, and does not have to compute the state sensitivities with respect to the vector of parameters. Thus, solving the ODE is faster since no additional equations accounting for the state sensitivities are added. There are several effects of this second change. First, no term involving the residual function $\tilde{r}$ will appear in (\ref{eq:07191543}), since $\tilde{r}(\bm{x}_k,\theta_k) = 0$. Second, the residual function evaluated at $(\bm{x}_{k+1},\theta_k)$ becomes
$$\tilde{r}(\bm{x}_{k+1},\theta_k) = -\alpha_{\bm{x}}\nabla_{\bm{x}}\tilde{L}(\bm{x}_k),$$
while the Jacobian $\frac{\partial \tilde{r}}{\partial \theta}(\bm{x}_{k+1}, \theta_k)$ remains unchanged since it does not depend on $\bm{x}$, i.e.,
$$\frac{\partial \tilde{r}}{\partial \theta}(\bm{x}_{k+1}, \theta_k) = \frac{\partial \hat{\bm{x}}}{\partial \theta}(\theta_k).$$
Therefore, we obtain the new iterations
\begin{eqnarray}
\label{eq:07191553}
\bm{x}_{k} &=& \textmd{\tt ODESolver}(\theta_k;x_0),\\
%\label{eq:07191554}
%  \bm{x}_{k+1} &=& \bm{x}_k - \nabla_{\bm{x}}L(\bm{x}_k),\\
\label{eq:07191555}
 \theta_{k+1} &=& \theta_k - \alpha\left[\left(\frac{\partial \hat{\bm{x}}}{\partial \theta}(\theta_k)\right)^T\nabla_{\bm{x}}\tilde{L}(\bm{x}_k)\right].
\end{eqnarray}
where $\alpha = \alpha_{\theta}\alpha_{\bm{x}}$ and (\ref{eq:07191555}) is exactly the iteration (\ref{eq:07201657}). Hence, unsurprisingly, we have recovered the gradient descent algorithm for solving (\ref{eq:07182001}). It should be clear by now that to avoid having to explicitly compute the sensitivities of the state vector with respect to $\theta$, we change the residual function $\tilde{r}$ by another residual function that does not have such requirements. Such a residual function was defined in (\ref{eq:07202115}) and its evaluation is efficient since batches of time instances can be executed in parallel.
Assuming Lipschitz continuity of $f(x)$, we are guaranteed by the Picard–Lindelöf theorem \cite{Lindelof_1894aa} that the ODE has a solution and it is unique. Hence by replacing the residual function $\tilde{r}(\bm{x},\theta)$ with ${r}(\bm{x},\theta)$, we do not change the solution of (\ref{eq:071921301}).
Retracing the previous steps, where instead of using the residual function $\tilde{r}$, we use $r$, we recover {Algorithm \ref{alg2}}.
We note that since $\bm{x}_{k}$ is a solution of the ODE, ${r}(\bm{x}_{k},\theta_k)=0$, and therefore $\nabla_{\bm{x}}F(\bm{x}_{k},\theta_{k}) = \nabla_{\bm{x}}\tilde{L}(\bm{x}_{k})$. The gradients and Jacobians that appear in  {Algorithm \ref{alg1}} and {Algorithm \ref{alg2}} can be computed using automatic differentiation.

\subsection{Enforcing the residual function equality constraints}
If $\tilde{L}$ does not attain its minimum at zero, we need to make sure that the norm of the residual function is forced to be as small as possible. We can achieved this by introducing a hyper-parameter $\lambda$ that acts as a weight for the residual function. The new cost function becomes $F(\bm{x},\theta)  = \tilde{L}(\bm{x})+\frac{\lambda}{2}\|r(\bm{x},\theta)\|^2$ and we can do a hyper-parameter search to reduce the magnitude of the residual function. A better approach is to update $\lambda$ during the optimization process, based on the magnitude of residual function. A guided strategy for updating online the weight function is based on a variant of the Augmented Lagrangian method \cite{Hestenes_1969} that employs the extended Lagragian function:
$\mathcal{L}_{\rho}(\bm{x}, \theta, \lambda) = L(\bm{x}) + \lambda^T r(\bm{x}, \theta)+\frac{\rho}{2}\|r(\bm{x},\theta)\|^2$, for $\rho>0$. In this method, we solve a min-max optimization problem: $\min_{\bm{x}, \theta}\max_{\lambda}\mathcal{L}(\bm{x}, \theta, \lambda)$. An extension of Algorithm \ref{alg1} that enforces the residual equality constraint is summarized in Algorithm \ref{alg3}.

\begin{algorithm}
\caption{ Augmented Lagrangian method based on gradient descent iterations }
\label{alg3}
\begin{algorithmic}
\REQUIRE $\alpha_x$, $\alpha_{\theta}$: Stepsizes
\REQUIRE $r(\bm{x},\theta)$: Residual function as implicit dynamical constraints
\REQUIRE $x_0$: Initial state vector\\
\REQUIRE $\rho>0$: Residual function norm weight\\
\REQUIRE $\theta_0$: Initial parameter vector\\
        $\lambda_0$: Initial Lagrange multiplier vector\\
        $\bm{x}_0 \leftarrow \textmd{\tt ODESolver}(\theta_0;x_0)$\\
        $k \leftarrow 0$
\WHILE{$\theta_k$, $\bm{x}_k$, $\lambda_k$ not converge}
\STATE $k\leftarrow k+1$
\STATE $\bm{x}_k \leftarrow \bm{x}_{k-1} -\alpha_x \nabla_{\bm{x}}\mathcal{L}_{\rho}(\bm{x}_{k-1},\theta_{k-1}, \lambda_{k-1})$
\STATE $\theta_k \leftarrow \theta_{k-1} -\alpha_{\theta} \nabla_{\theta}\mathcal{L}_{\rho}(\bm{x}_k,\theta_{k-1}, \lambda_{k-1})$
\STATE $\lambda_k \leftarrow \lambda_{k-1} + \rho r(\bm{x}_k,\theta_{k})$
\ENDWHILE
\RETURN $\bm{x}_k$,  $\theta_k$
\end{algorithmic}
\end{algorithm}

\section{Experiments}
We tested Algorithms \ref{alg0} - \ref{alg2} on the Cucker-Smale model describing the interaction between particles that include self-propelling, friction and attraction-repulsion phenomena. It takes into account an alignment mechanism of the particles by averaging their relative velocities with all the other particles. The strength of this averaging process depends on the mutual distance. For $N$ particles the model is described by the following dynamical system:
\begin{eqnarray}
  \dot{x}_i &=& v_i, \\
  \dot{v}_i &=& \frac{1}{N}\sum_{j=1}^N H(\|x_i-x_j\|)(v_j - v_i) - \frac{1}{N}\nabla U(\|x_i-x_j\|),
\end{eqnarray}
for $i=1,\ldots, N$, where $x_i$ and $v_i$ are the two-dimensional particle positions and velocities, $H(r)=\frac{1}{(1+r^2)^{\gamma}}$ is the communication rate, and  $U(r)=-c_ae^{-r/l_a}+c_re^{-r/l_r}$ is the inter particle potential energy. The parameter $\gamma$ is a positive scalar, and $c_a$, $c_r$ and $l_a$, $l_r$ are the strength and the length of the attraction and repulsion, respectively. The number of states is linear in the number of particles, i.e., $4N$.
The learning problem is defined as follows: given the particle trajectories (i.e., time series of positions and velocities) learn the parameters of model.

 We tested the algorithms for various number of particles: $N\in\{5, 10, 20, 50, 100, 200\}$. We used one time series as training data, generated with the same, random, initial parameters, for all cases. The training loss function is the sum of squared errors (SSE). The test data consists of model trajectories generated with random initial conditions. We compared Algorithms \ref{alg0} - \ref{alg2} using four metrics: total optimization time, training loss function (i.e., SSE), training relative sum of squared error (RSSE) based on ODE solutions, and  RSSE on test data. We run the algorithm for 5k epochs and evaluated the four metrics for all combinations of algorithm and number of particles. The RSSE metric is computed with respect to the solution of the ODE for a particular instance of the model parameters generated during the optimization process. This way we can compare how far the solution of the ODE is from the target trajectories. The total optimization time is computed as the product between the average epoch time and the number of epochs. Since Algorithms \ref{alg0} and \ref{alg2} must solve ODEs, their epochs are computationally more expensive. We implemented the three algorithms on both Pytorch and Jax and used their ODE solvers to generate ODE solutions and state sensitivities, when needed. In particular, we used the adaptive step {\tt Dopri5} ODE solver. The Jax implementation is faster since we took advantage of  the just in time compiling (JIT) feature.

\subsection{Pytorch implementation}
The Pytorch implementation details of the three algorithms are shown in Table \ref{tab:algorithm implementation}.
\begin{table}[ht!]
\caption{Algorithms \ref{alg0}-\ref{alg2} Pytorch implementation details.}
\label{tab:algorithm implementation}
\centering
\begin{tabular}{|c|c|c|c|}
\hline
 & \textbf{Sate update} & \textbf{Parameter update} & ODE solver\\
\hline
\textbf{Algorithm \ref{alg0}} & X & Adam ($l_r=0.01$) & Dopri5\\
\hline
\textbf{Algorithm \ref{alg1}} & SGD ($l_r=0.01$) & Adam ($l_r=0.01$) & X\\
\hline
\textbf{Algorithm \ref{alg2}} & SGD ($l_r=1$) & Adam ($l_r=0.01$) & Dopri5\\
\hline
\end{tabular}
\vspace{-8pt}
\end{table}
In the case of Algorithm \ref{alg1} we used SGD and Adam algorithms for updating the state and the parameters, respectively. Both algorithm used a stepsize $l_r=0.01$. In the case of Algorithm \ref{alg1}, we used the same combination of algorithms, but the stepsize for SGD is $l_r=1$. All experiments were executed on a GPU device.
We used the ODE solver in the {\tt torchdiffeq} library for Algorithm \ref{alg0} and \ref{alg2}. In the case of Algorithm \ref{alg0} we tested the performance of both direct and adjoint methods for the sensitivity analysis. Both options are provide by the {\tt torchdiffeq} library. In the case of Algorithm \ref{alg2}, the ODE solver was used to compute the ODE solution only, and not the state sensitivities.  Table \ref{tab:Pytorch_epoch_time} shows the epoch time for each of the three algorithms for various number of particles. As expected, Algorithm \ref{alg2} average iteration time is smaller than Algorithm \ref{alg0} iteration time, since no sensitivities are computed.
%This difference between these values will increase as the number of parameters increases, since the size of the ODE corresponding to the state sensitivity dynamics increases, as well.
Algorithm \ref{alg1} is at least 4x faster then Algorithms \ref{alg0} since it makes the best use of the parallel computations enabled by the GPU.  GPU was used for all experiments related to Algorithm \ref{alg0} resulting in rather flat average iteration time. Slight improvements in time per epoch can be obtain if the CPU is used for small number of particles. However, CPU usage does not scale with the number of particles. In addition, not unexpectedly, the implementation that uses the adjoint method is much slower since the number of state variables dominate the number of parameters.
\begin{table}[ht!]
\caption{Pytorch: Algorithms \ref{alg0}-\ref{alg2} average time per epoch in seconds. Improvements of Algorithms 2 and 3 over the fastest version of Algorithm 1 are shown in parentheses.}
\label{tab:Pytorch_epoch_time}
\centering
\resizebox{\columnwidth}{!}{%
\begin{tabular}{|c|c|c|c|c|c|c|}
\hline
 & \textbf{N=5} & \textbf{N=10} & \textbf{N=20} & \textbf{N=50} & \textbf{N=100} & \textbf{N=200}\\
\hline
\textbf{Algorithm \ref{alg0}, direct} & 0.463 & 0.602 & 0.594 & 0.565 & 0.606 & 0.614\\
\hline
\textbf{Algorithm \ref{alg0}, adjoint} & 2.655 & 2.651 & 2.665 & 2.673 & 2.681 & 2.819\\
\hline
\textbf{Algorithm \ref{alg1}} & 0.009 (50x) & 0.008 (75x) & 0.008 (74x) & 0.011 (51x) & 0.037 (16x) & 0.155 (4x)\\
\hline
\textbf{Algorithm \ref{alg2}} & 0.171 (3x) & 0.188 (3x) & 0.192 (3x) & 0.179 (3x) &0.184 (3x) & 0.279 (2x)\\
\hline
\end{tabular}
}
\vspace{-8pt}
\end{table}

Figure \ref{fig:SSE_pytorch} shows the training loss for the three algorithms in logarithmic scale. Only the training losses for Algorithms \ref{alg1} and \ref{alg2} are truly comparable since they have the same loss function. We note that  Algorithm  \ref{alg0}'s loss function at the end of training is comparable to that of Algorithm \ref{alg2}. The results show that Algorithm \ref{alg1} has the best performance in training, measured by the SSE value and the time it needed to reach the SSE value, and tt is followed by Algorithm \ref{alg2}.  Algorithm \ref{alg0} has the worst performance in training, however since the loss function is defined with respect to a different residual function, we cannot draw a definitive conclusion.
\begin{figure}[ht!]
\includegraphics[width=\textwidth,clip]{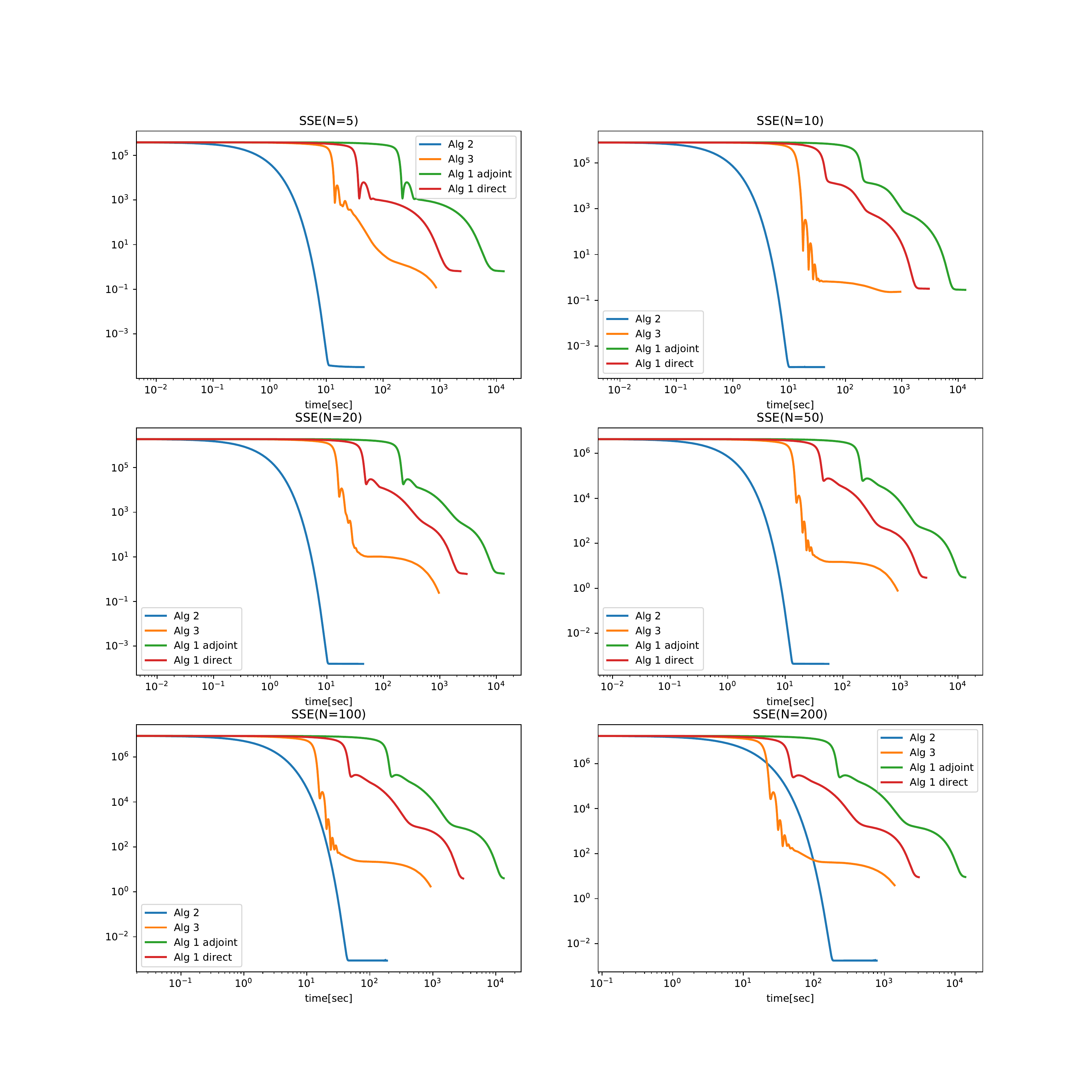}
\caption{Pytorch: SSE training loss comparison for various number of particles.}
\label{fig:SSE_pytorch}
\end{figure}
More insights on the performance of the three algorithms come from comparing the RSSE plots. The RSSE metrics are computed based on the ODE solutions at the model parameter instances generated during the optimization process. Figure \ref{fig:RSSE_pytorch} shows the RSSE plots of the three algorithms for various numbers of particles. All observations we made from the SSE plots carry to this set of plots. Algorithms \ref{alg1} and \ref{alg2} are superior to Algorithm \ref{alg0} (both versions) in  convergence time and best achieved RSSE value.
\begin{figure}[ht!]
\includegraphics[width=\textwidth,clip]{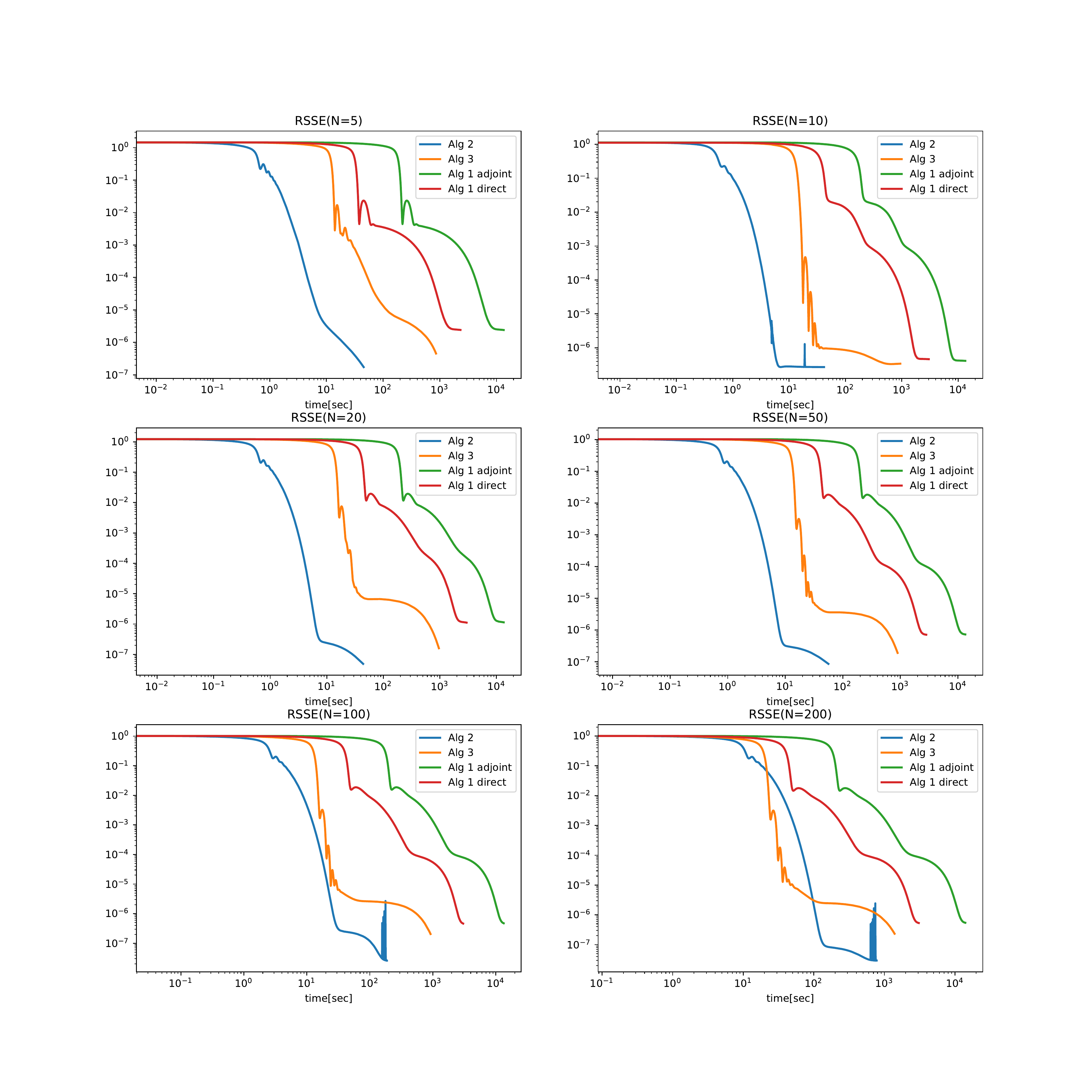}
\caption{Pytorch: RSSE training loss comparison for various number of particles.}
\label{fig:RSSE_pytorch}
\end{figure}
We evaluated the accuracy of the  results against test data generated using 100 randomly selected initial conditions. We computed the average RSSE metrics for each case, and reported the results in Table \ref{tab:Pytorch_test_results}. Similar to the training results, Algorithm \ref{alg1} gives the most accurate results. It is followed by Algorithm \ref{alg2} and Algorithm \ref{alg0}.

\begin{table}[ht!]
\caption{Pytorch: Algorithms \ref{alg0}-\ref{alg2} RSSE on test data.}
\label{tab:Pytorch_test_results}
\centering
\resizebox{\columnwidth}{!}{%
\begin{tabular}{|c|c|c|c|c|c|c|}
\hline
 & \textbf{N=5} & \textbf{N=10} & \textbf{N=20} & \textbf{N=50} & \textbf{N=100} & \textbf{N=200}\\
\hline
\textbf{Algorithm \ref{alg0}, direct} & $3.97\times 10^{-5}$ & $2.42\times 10^{-6}$ & $1.82\times 10^{-6}$ & $7.26\times 10^{-7}$ & $5.00\times 10^{-7}$ & $4.32\times 10^{-7}$\\
\hline
\textbf{Algorithm \ref{alg0}, adjoint} & $3.98\times 10^{-5}$ & $2.44\times 10^{-6}$ & $1.85\times 10^{-6}$ & $7.40\times 10^{-7}$ & $5.12\times 10^{-7}$ & $4.41\times 10^{-7}$\\
\hline
\textbf{Algorithm \ref{alg1}} & $1.65\times 10^{-6}$ & $2.50\times 10^{-7}$ & $1.77\times 10^{-7}$ & $8.52\times 10^{-8}$ & $3.70\times 10^{-8}$ & $3.24\times 10^{-8}$\\
\hline
\textbf{Algorithm \ref{alg2}} & $3.65\times 10^{-6}$ & $1.85\times 10^{-7}$ & $3.18\times 10^{-7}$ & $1.96\times 10^{-6}$ & $2.37\times 10^{-7}$ & $1.98\times 10^{-7}$\\
\hline
\end{tabular}
}
\vspace{-8pt}
\end{table}

\subsection{Jax implementation}
We repeated the experiments using Jax implementations. One advantage  using Jax is the speedups enabled by JIT. The optimization algorithms implementation details are shown in Table \ref{tab:algorithm implementation jax}. In the case of Algorithm \ref{alg1}, for stability reasons, we changed the stepsizes for the SGD and Adam algorithm, while making sure that their product is equal to the stepize product used in the Pytorch implementation.
\begin{table}[ht!]
\caption{Algorithms \ref{alg0}-\ref{alg2} Jax implementation details.}
\label{tab:algorithm implementation jax}
\centering
\begin{tabular}{|c|c|c|c|}
\hline
 & \textbf{Sate update} & \textbf{Parameter update} & ODE solver\\
\hline
\textbf{Algorithm \ref{alg0}} & X & Adam ($l_r=0.01$) & Dopri5\\
\hline
\textbf{Algorithm \ref{alg1}} & SGD ($l_r=0.01$) & Adam ($l_r=0.01$) & X\\
\hline
\textbf{Algorithm \ref{alg2}} & SGD ($l_r=0.1$) & Adam ($l_r=0.1$) & Dopri5\\
\hline
\end{tabular}
\vspace{-8pt}
\end{table}
The times per iteration in the Jax implementation are shown in Table \ref{tab:Jax_epoch_time}. Similar to the Pythorch implementation, Algorithms \ref{alg1} and \ref{alg2} are much faster than Algorithm \ref{alg0}. In addition, the training losses for Algorithms \ref{alg1} and \ref{alg2} are consistently smaller than Algorithm \ref{alg0}'s training loss. Algorithm \ref{alg0} uses the Jax provided ODE solver with the same {\tt Dopri5} integration scheme, as in the Pytorch case. The sensitivity analysis in the ODE solver implementation is based on the adjoint method. We note even when using JIT, the time per iteration of the Jax implementation of Algorithm \ref{alg0} is significantly slower than the Pytorch implementation using the direct method, but much faster that the adjoint version. In the case of Algorithms \ref{alg1} and \ref{alg2}, the time efficiency is dramatically better in Jax than in Pytorch, mainly due to using the JIT feature.
\begin{table}[ht!]
\caption{Jax: Algorithms \ref{alg0}-\ref{alg2} average time per epoch in seconds. Improvements of Algorithms 2 and 3 over Algorithm 1 are shown in parentheses.}
\label{tab:Jax_epoch_time}
\centering
\resizebox{\columnwidth}{!}{%
\begin{tabular}{|c|c|c|c|c|c|c|}
\hline
 & \textbf{N=5} & \textbf{N=10} & \textbf{N=20} & \textbf{N=50} & \textbf{N=100} & \textbf{N=200}\\
\hline
\textbf{Algorithm \ref{alg0}, adjoint} & 0.935 & 0.973 & 1.243 & 1.526 & 1.728 & 1.896\\
\hline
\textbf{Algorithm \ref{alg1}} & 0.0012 (x780) & 0.0014 (x695) & 0.0015 (x827) & 0.0013 (x1173) & 0.0014 (x1234) & 0.0013 (x1458)\\
\hline
\textbf{Algorithm \ref{alg2}} & 0.06 (x16) & 0.06 (x16) & 0.061 (x20) & 0.062 (x25) & 0.065 (x27) & 0.072 (x27)\\
\hline
\end{tabular}
}
\vspace{-8pt}
\end{table}
The training losses SSE and RSSE are shown in Figures \ref{fig:SSE_jax} and \ref{fig:RSSE_jax}, respectively.
\begin{figure}[ht!]
\includegraphics[width=\textwidth,clip]{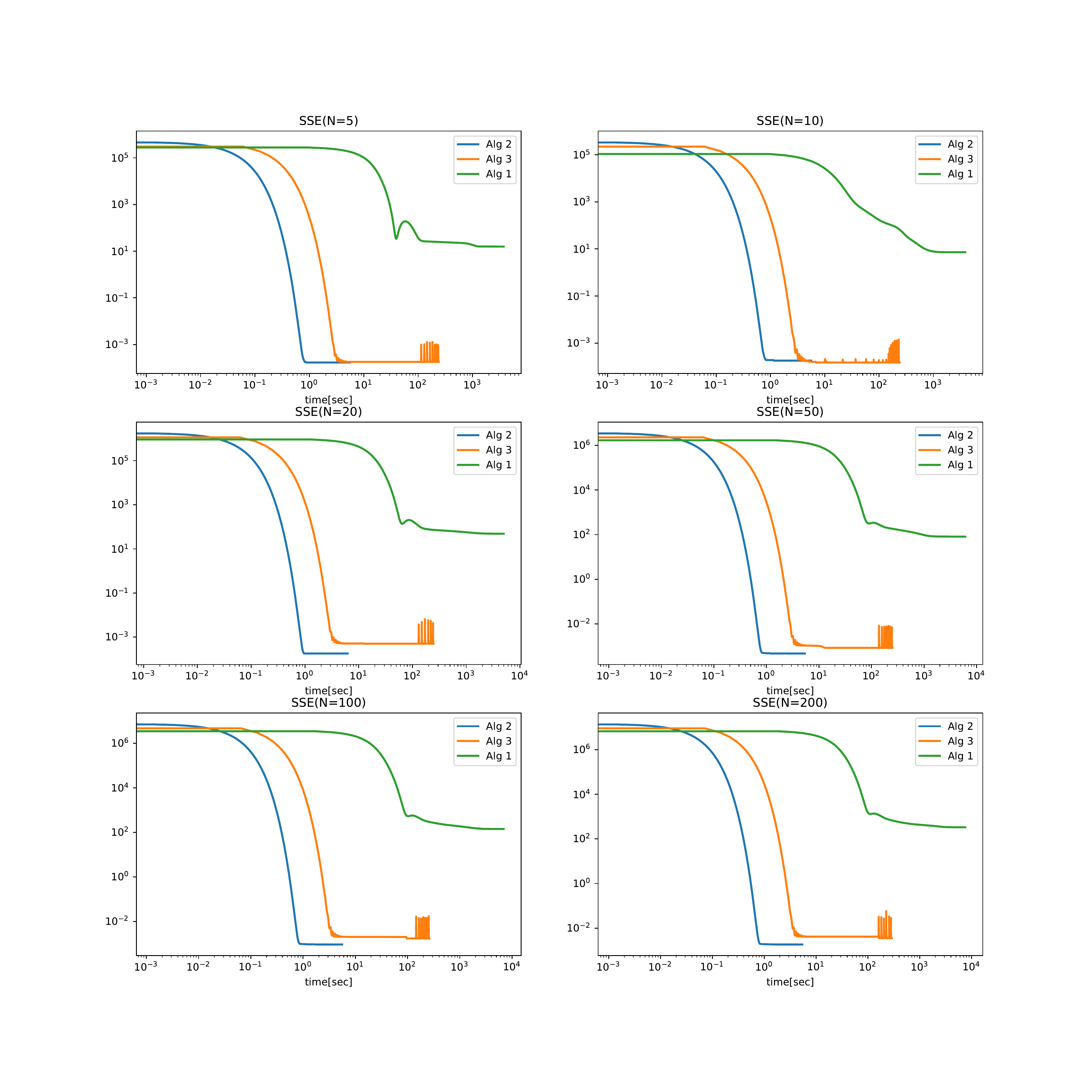}
\caption{Jax: SSE training loss comparison for various number of particles.}
\label{fig:SSE_jax}
\end{figure}
\begin{figure}[ht!]
\includegraphics[width=\textwidth,clip]{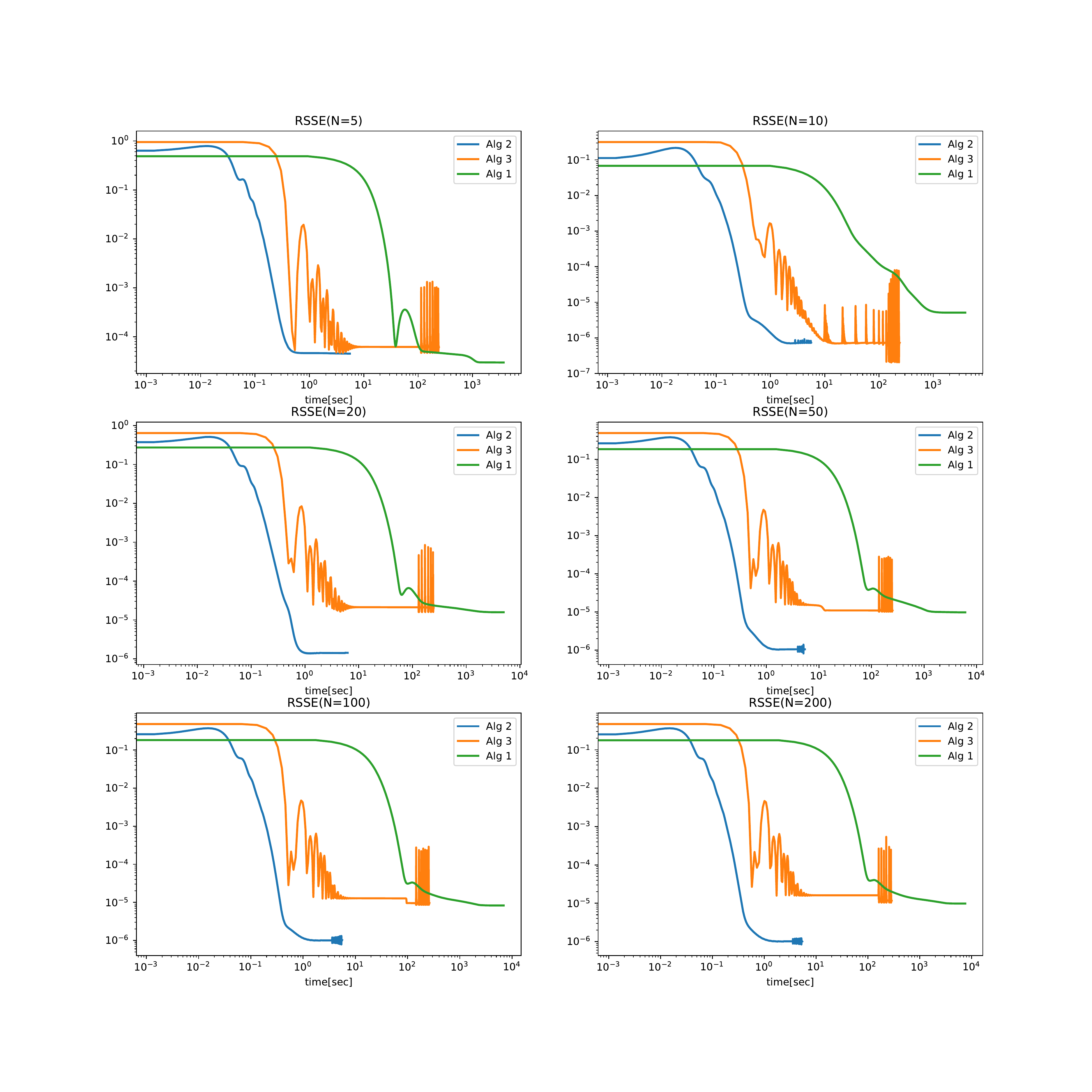}
\caption{Jax: RSSE training loss comparison for various number of particles.}
\label{fig:RSSE_jax}
\end{figure}
We run the models learned using Algorithms \ref{alg0} - \ref{alg2} on test data, on the same test data used in the Pytorch case. The results are shown in \ref{tab:Jax_test_results}, showing that Algorithm \ref{alg1} fares better than both Algorithms \ref{alg0} and \ref{alg2}, while Algorithm \ref{alg2} is comparable in accuracy to Algorithm \ref{alg0}.

\begin{table}[ht!]
\caption{Jax: Algorithms \ref{alg0}-\ref{alg2} RSSE on test data.}
\label{tab:Jax_test_results}
\centering
\resizebox{\columnwidth}{!}{%
\begin{tabular}{|c|c|c|c|c|c|c|}
\hline
 & \textbf{N=5} & \textbf{N=10} & \textbf{N=20} & \textbf{N=50} & \textbf{N=100} & \textbf{N=200}\\
\hline
\textbf{Algorithm \ref{alg0}} & $9.78\times 10^{-5}$ & $5.31\times 10^{-5}$ & $2.43\times 10^{-5}$ & $2.10\times 10^{-5}$ & $1.94\times 10^{-5}$ & $1.94\times 10^{-5}$\\
\hline
\textbf{Algorithm \ref{alg1}} & $7.78\times 10^{-5}$ & $6.74\times 10^{-7}$ & $1.86\times 10^{-7}$ & $1.40\times 10^{-7}$ & $1.56\times 10^{-7}$ & $1.46\times 10^{-7}$\\
\hline
\textbf{Algorithm \ref{alg2}} & $9.52\times 10^{-5}$ & $2.29\times 10^{-6}$ & $3.02\times 10^{-5}$ & $2.43\times 10^{-5}$ & $2.25\times 10^{-5}$ & $2.17\times 10^{-5}$\\
\hline
\end{tabular}
}
\vspace{-8pt}
\end{table}

\section{Related work}

The authors of \cite{NEURIPS2018_69386f6b} introduced the notion of ``neural'' ODE, where they parameterized the right-hand side $f$ of an ODE by a neural network. The proposed approach for computing gradients of loss functions used the adjoint method for implementing sensitivity analysis, and avoids using backpropagation. Thus, it eliminates the memory cost and numerical errors induced by differentiating through the operations of the forward pass.   The authors of \cite{NEURIPS2018_69386f6b} made available a Pytorch library named {\tt torchdiffeq} that enables compositions of neural-ODEs with Pytorch layers. Moreover, the parameters of the resulting model can be trained simultaneously. The library {\tt torchdiffeq} includes both direct and adjoint methods to support sensitivity analysis. In \cite{NEURIPS2018_69386f6b,NEURIPS2019_42a6845a,grathwohl2019ffjord}, the authors use neural ODEs to learn latent time series models, density models, or as a replacement for very deep neural networks. These algorithms are closely related to the sensitivity analysis methods introduced in \cite{gardner2022sundials,hindmarsh2005sundials}.  They include explicit ODE solvers only since they do not require Newton-Raphson steps to solve the implicit nonlinear equations. Our approach to integrating dynamical constraints described in Algorithms \ref{alg1} and \ref{alg2} is compatible with Pytorch models as well. Parameter training would require the adding a loss function for minimizing the ODE-induced residuals. When using the state-reset approach, we can use any ODE solver, implicit or explicit. Jax \cite{jax2018github} includes the ability to solve ODEs as well, and it does include the {\tt Dopri5} solver we used for comparison purposes. In the current implementation, the state sensitivities are computed using adjoint methods only. The Jax based ODE solver library {\tt Diffrax} \cite{kidger2021on} includes both direct and adjoint methods for computing state sensitive, although the direct methods use forward time. This library includes a similar number of integrators as {\tt torchdiffeq}. One of the advantages of Jax was the ability to use the JIT decorator that results in tremendous computational efficiency. There are efforts to provides similar capability to Pytorch, however these efforts have not yet resulted in he same level of maturity as in the Jax case.
There is previous work on carrying differentiation operators over the steps of ODE solvers. Adjoint methods applied to both ODEs and PDEs together with backpropagation over the forward steps of the ODE/PDE solvers are implemented in the {\tt dolfin} library \cite{Farrell2013AutomatedDO}. Gradient estimations using direct sensitivity analysis methods was demonstrated in \cite{DBLP:journals/corr/CarpenterHBLLB15}. Choosing the direct of adjoint method depends on the problem. When the number of optimization variables dominates the size of the state vector, the adjoint methods are preferred. In contrast, when the number of optimization variables is small compared to the size of the state vector, direct methods are numerically more efficient. This was clear in the Cucker-Smale example, where the number of parameter remained constant, while the state vector dimension was increasing.
%this phenomenon was evident in the Pytorch vs. Jax algorithm implementations.
An alternative to direct collocation methods is the global parameterization of solutions approach \cite{doi:10.1073/pnas.1718942115}. Spectral methods based on representing differential equations solutions as expansion of basis functions (e.g., Chebyshev, Fourier) have a long history \cite{boyd01}. More recent approaches use neural networks to parameterize solutions of PDEs, where automatic differentiation is used the construct a loss function in terms of spatial and temporal differential operators.
Closer to our idea is the result shown in \cite{RERCSM21}, where the authors avoid using ODE solvers by approximating the state derivative using a collocation method. In their approach, the loss function is defined in terms of the approximation of the observed state derivative and the one predicted by the a neural-ODE. The full state needs to be observed. Our implementation apply to partially observed systems, as well. In addition, we improve the speed of convergence via a coordinated descent approach.

%\section{Conclusion}
%We hope that these files are useful to you in preparing your submission for IEEE CSM. We again remind you to carefully study the IEEE CSM Author’s Guide. If you have any questions, do not hesitate to contact the Editor-in-Chief Jonathan P. How at jhow@mit.edu.

\bibliographystyle{IEEEtran}
\bibliography{references,refs}

\newpage

\end{document}